\documentclass{tlp}

\usepackage{times}
\usepackage{amsmath}
\usepackage{graphicx}
\usepackage{multirow}
\usepackage[hidelinks]{hyperref}
\usepackage[utf8]{inputenc}
\usepackage[small]{caption}
\usepackage{graphicx}
\usepackage{booktabs}
\usepackage[linesnumbered,ruled,vlined]{algorithm2e}
\usepackage{fancyvrb}
\usepackage{xspace}
\usepackage{subcaption}
\usepackage{pgfgantt}
\usepackage{microtype}
\usepackage{mathtools}
\usepackage{rotating}
\usepackage{footmisc}

\newcommand{\atom}[2]{\ensuremath{\mathit{#1}_{#2}}}

\newtheorem{definition}{Definition}
\newtheorem{proposition}{Proposition}

\newcommand{\mcap}[1]{\mathit{cap}({#1})}
\newcommand{\mspan}[1]{\mathit{span}({#1})}

\newcommand{\solve}[1]{\mathit{solve}({#1})}
\newcommand{\opt}[1]{\mathit{opt}({#1})}
\newcommand{\bound}[1]{\mathit{bound}({#1})}

\newcommand{\clingo}{\texttt{clingo}\xspace}
\newcommand{\clingoDL}{\texttt{clingo-dl}\xspace}
\newcommand{\cplex}{\texttt{cplex}\xspace}

\newcommand{\ortools}{\texttt{or-tools}\xspace}
\newcommand{\gecode}{\texttt{gecode}\xspace}
\newcommand{\cpoptimizer}{\texttt{cpoptimizer}\xspace}
\newcommand{\clingcon}{\texttt{clingcon}\xspace}

\begin{document}

%\lefttitle{Thomas Eiter, Tobias Geibinger, Nysret Musliu, Johannes Oetsch, Peter Skocovsky, Daria Stepanova}
\lefttitle{T.~Eiter, T.~Geibinger, N.~Musliu, J.~Oetsch, P.~Skočovský, and D.~Stepanova}

\jnlPage{1}{25}
\jnlDoiYr{2022}
\doival{10.1017/xxxxx}

\title[ASP for Lexicographical Makespan Optimisation in Parallel Machine Scheduling]{Answer-Set Programming for  Lexicographical Makespan Optimisation in Parallel\\ Machine Scheduling\thanks{
This is an extended version of a paper \citep{eiter2021answer} that appeared in the Proceedings of the 18th International Conference on Principles of
Knowledge Representation and Reasoning (KR’21).
}}

\begin{authgrp}
\author{\sn{Thomas} \gn{Eiter}, \sn{Tobias} \gn{Geibinger}, \sn{Nysret} \gn{Musliu}, \sn{Johannes} \gn{Oetsch}}
\affiliation{Institute of Logic and Computation,\\ 
Vienna University of Technology \textup{(}TU Wien\textup{)}, Austria \\
\emailsfour{eiter@kr.tuwien.ac.at}{tgeibing@dbai.tuwien.ac.at}{musliu@dbai.tuwien.ac.at} {oetsch@kr.tuwien.ac.at}}
\author{\sn{Peter} \gn{Skočovský}, \sn{Daria} \gn{Stepanova}}
\affiliation{Bosch Center for AI,\\
    Robert Bosch Campus 1, D-71272 Renningen, Germany \\
\emailstwo{fixed-term.peter.skocovsky@de.bosch.com}{daria.stepanova@de.bosch.com}}
\end{authgrp}

%\history{\sub{xx xx xxxx;} \rev{xx xx xxxx;} \acc{xx xx xxxx}}

\maketitle

\begin{abstract}
We deal with a challenging scheduling problem on parallel machines with sequence-dependent setup times and release dates from a real-world application of semiconductor work-shop production. There, jobs can only be processed by dedicated machines, thus few machines can determine the makespan almost regardless of how jobs are scheduled on the remaining ones. This causes problems when machines fail and jobs need to be rescheduled. Instead of optimising only the makespan, we put the individual machine spans in non-ascending order and lexicographically minimise the resulting tuples. This achieves  that all machines complete as early as possible and increases the robustness of the schedule. We study the application of Answer-Set Programming (ASP) to solve this problem. While ASP eases modelling, the combination of timing constraints and the considered objective function challenges current solving technology. The former issue is addressed by using an extension of ASP by difference logic. For the latter, we devise different algorithms that use multi-shot solving. To tackle industrial-sized instances, we study different approximations and heuristics. Our experimental results show that ASP is indeed a promising KRR paradigm for this problem and is competitive with state-of-the-art CP and MIP solvers.
Under consideration in Theory and Practice of Logic Programming (TPLP).
\end{abstract}


%%%%%%%%%%%%%%%%%%%%%%%%%%%%%%%%%%%%%%%%%%%%%%%%%%%%%%%%%%%%%%%%%%%%%%%%%%%%%%%%%%%%%
\section{Introduction}\label{sec:intro}
%%%%%%%%%%%%%%%%%%%%%%%%%%%%%%%%%%%%%%%%%%%%%%%%%%%%%%%%%%%%%%%%%%%%%%%%%%%%%%%%%%%%%

We  consider a scheduling problem on unrelated parallel machines 
that arises in industrial semiconductor production at Bosch. 
The problem is  involved due to several aspects. We have to deal with sequence-dependent setup-times and job release dates on the one hand; on the other hand, setup-times, release dates, and job durations are machine dependent, and jobs can only be scheduled on dedicated machines.
Our goal is to maximise the \emph{throughput}, which can be defined as the number of jobs processed per time unit. 
In principle, minimising the \emph{makespan}, i.e., the total length of the schedule, accomplishes this.
However, we have to deal with \emph{high machine dedication}, i.e., many jobs can be processed only by few machines. 
When dealing with machines with high dedication, 
few machines determine the makespan almost regardless of how jobs are scheduled on the remaining ones.
This is not ideal when jobs need to be rescheduled because of, e.g., machine failure, and
domain experts expressed the requirement that 
\emph{``all machines should complete as early as possible''} to
give the scheduler freedom for rearrangements. 

Instead of optimising only the makespan, we lexicographically minimise the individual machine spans. In particular, we define the \emph{lexicographical makespan} of a schedule as the tuple of all machine spans in non-ascending order. We prefer a schedule with smaller lexicographical makespan over one with a larger one, where we use lexicographical order for comparison.
A schedule with minimal lexicographical makespan has therefore also a minimal makespan, but ties are broken using machines that complete earlier.

This idea of \emph{lexicographical optimisation} to produce schedules that
show to be robust when they need to be updated has been investigated and confirmed
in the context of job scheduling on identical machines in recent work~\citep{letsios2021exact}.
While these results further support our motivation to use this objective, the algorithms and tool chains developed there cannot be used directly as our problem is significantly more complex due to machine dedication, 
sequence-dependent setup times, and release dates.  

We are specifically interested in using \emph{Answer-Set Programming} (ASP)~\citep{brewka2011answer,gebser2012answer,lifschitz2019answer}, a state-of-the-art logic-based KRR paradigm, for our scheduling problem. 
ASP is interesting for two reasons: first, its expressive modelling language makes it easy to 
concisely model the problem including the objective function.
This allows one to quickly come up with a first prototype that can 
be evaluated by domain experts and 
can serve as a blueprint for more sophisticated solutions.
Second and more importantly, ASP makes it relatively easy 
to develop solutions which can be 
conveniently adapted to problem variations,  a feature known as \emph{elaboration tolerance}~\citep{mccarthy1998elaboration}. This is indeed needed as it is a goal to 
use similar scheduling solutions for other work centers with slightly different requirements.

While ASP makes modelling easy and provides enough flexibility for future adaptations, the combination of timing constraints and the considered objective function challenges current solving technology. The former issue is addressed by using an extension of ASP with difference logic~\citep{janhunen2017clingo}. 
While this solves the issue of dealing with integer domains without blowing up the size of the grounding, it makes expressing optimisation more tricky as current technology does not support complex optimisation of integer variables.
We devise different algorithms that use multi-shot solving~\citep{gebser2019multi} to 
accomplish lexicographical optimisation for our scheduling problem. 

To tackle industrial-sized instances, we study different approximations and heuristics.
In particular, we consider an approximate algorithm where parts of a solution are fixed after solver calls.  This allows us to find near-optimal solutions in a short time.  
Orthogonally 
to the ASP model, we formulate different \emph{heuristic rules} using a respective ASP extension~\citep{gebser2013domain}. These rules do not alter the solution space but guide the solver with variable assignments and  
improve performance.

For an experimental evaluation of our algorithms,
we use random instances of various sizes that are generated based on real-life scenarios. In addition,  we provide an alternative solver-independent MiniZinc model that can be used by various state-of-the-art MIP and CP solvers.
The experiments
aim to explore 
the additional costs needed when using the lexicographical makespan for optimisation instead of the standard makespan and 
the trade-off between performance and solution quality. 
Our experimental results show that 
the lexicographical makespan optimisers produce schedules with small makespans and thus ensure high throughput, while at the same time accomplish our 
other objective of early completion times for all machines.
The ASP-based algorithms scale up 
to instances of realistic size and demonstrate 
that ASP is indeed a viable KRR solving paradigm for lexicographical makespan problems.

This article is an extended version of a conference paper that has been presented at KR~2021 \citep{eiter2021answer}. In particular, it includes proofs, further encodings and additional experimental data that were not contained in the original publication, as well as an extended discussion and other minor extensions.

The rest of this paper is organised as follows.
After some background on ASP in the next section,  
we formally define our scheduling problem including the lexicographical makespan objective in Section~\ref{sec:problem}. 
We then present 
exact ASP approaches for lexicographical makespan minimisation in Section~\ref{sec:exact}, and
discuss approximation approaches in Section~\ref{sec:approx}. 
Experiments are discussed 
in Section~\ref{sec:experiments}. We  review relevant literature in Section~\ref{sec:rel}, before we conclude in Section~\ref{sec:concl}.

%%%%%%%%%%%%%%%%%%%%%%%%%%%%%%%%%%%%%%%%%%%%%%%%%%%%%%%%%%%%%%%%%%%%%%%%%%%
\section{Background on ASP}\label{sec:background}
%%%%%%%%%%%%%%%%%%%%%%%%%%%%%%%%%%%%%%%%%%%%%%%%%%%%%%%%%%%%%%%%%%%%%%%%%%%

\emph{Answer-Set Programming} (ASP)~\citep{brewka2011answer,gebser2012answer,lifschitz2019answer} provides a declarative modelling language that allows one to succinctly represent search and optimisation problems, for which solutions can be computed using dedicated ASP solvers.\footnote{\url{potassco.org}.}\footnote{\url{www.dlvsystem.com}.}

ASP is a compact relational, in essence propositional, formalism where variables in the input language are replaced by constant symbols in a preprocessing step called \emph{grounding}.
An ASP program is a (finite) set of rules 
of the form
\begin{equation}%\small
\label{eq:rule}
\atom{p}{1}\, \mbox{\tt |}\, \ldots\, \mbox{\tt |}\, \atom{p}{k} \,\, \mbox{\tt :-} \,\, \atom{q}{1}, \ldots, \atom{q}{m},\, \mbox{\tt not}\, \atom{r}{1}, \ldots,\, \mbox{\tt not}\, \atom{r}{n} \mbox{\tt.}
\end{equation}
where all \atom{p}{i}, \atom{q}{j}, and \atom{r}{l} are function-free first-order atoms.%
\footnote{Extensions of ASP with strong negation as well as function symbols, both uninterpreted and interpreted ones, are available, which however we disregard here.}
The head 
are all atoms before the implication symbol $\mbox{\tt:-}$, and the body are all the atoms and negated atoms afterwards.
The intuitive meaning of the rule (\ref{eq:rule}) is that if all atoms 
$\atom{q}{1}, \ldots, \atom{q}{m}$
can be derived, and there is no evidence for any of the atoms 
$\atom{r}{1}, \ldots, \atom{r}{n}$
(i.e., the rule fires) then at least one of 
$\atom{p}{1}, \ldots, \atom{p}{k}$
has to be derived.

A rule with an empty body (i.e., $m=n=0$) is called a \emph{fact},
with {\tt:-} usually omitted. 
Facts are used to express knowledge that is unconditionally true. 
A rule with empty head  (i.e., $k=0$) is a \emph{constraint}.
The body of a constraint cannot be satisfied by any answer set and is used to prune away unwanted solution candidates.

The semantics of an ASP program $P$ is given in terms of particular Herbrand models of its grounding $grd_C(P) = \bigcup_{r\in P} grd_C(r)$, where $grd_C(r)$ consists of all ground (variable-free) rules $r\theta$ that result from $r$ by some substitution $\theta$ of the variables in $r$ with elements from the set $C$ of constants, which by default are the constants occurring in $P$. % (then $C$ is omitted). 
A \emph{(Herbrand) interpretation} $I$ is a subset of the set $HB_C(P)$ of all ground atoms $p(c_1,\ldots,c_l)$ with a predicate $p$ occurring in $P$ and constants $c_1,\ldots,c_l$ from $C$.  It is a \emph{(Herbrand) model} of $P$, if each rule $r$ in $grd_C(P)$ of form (\ref{eq:rule}) is satisfied, i.e., either (i) $\{ \atom{p}{1},\ldots \atom{p}{k} \} \cap I \neq \emptyset$ or (ii) $\{\atom{q}{1},\ldots, \atom{q}{m} \} \not \subseteq I$ or (iii) $\{ \atom{r}{1},\ldots \atom{r}{n} \} \cap I \neq \emptyset$. Then $I$ is an \emph{answer set}\/ of $P$, if $I$ is a $\subseteq$-minimal model of the reduct $P^I$ of $P$ by $I$, which is given by
$$P^I = \{ \atom{p}{1}\, \mbox{\tt |}\, \ldots\, \mbox{\tt |}\, \atom{p}{k} \,\, \mbox{\tt :-} \,\, \atom{q}{1}, \ldots, \atom{q}{m} \mid r 
\in P, \{ \atom{r}{1},\ldots \atom{r}{n} \} \cap I = \emptyset \};$$ 
intuitively, $I$ must be a $\subseteq$-minimal model of all rule instances of $P$ 
whose negative body is satisfied by $I$.

A common syntactic extension are choice rules of the form
\begin{equation*}%\small
i \, \mbox{\tt\{} \atom{p}{1}, \ldots, \atom{p}{k}  \mbox{\tt\}}\, j 
\,\, \mbox{\tt :-} \,\, \atom{q}{1}, \ldots, \atom{q}{m},\, \mbox{\tt not}\, \atom{r}{1}, \ldots,\, \mbox{\tt not}\, \atom{r}{n} \mbox{\tt.}
\end{equation*}
The meaning is that if 
the rule fires, then some subset $S$  of 
$\atom{p}{1}, \ldots, \atom{p}{k}$
with $i \leq |S| \leq j$  has to be true as well; it is compiled to ordinary rules using hidden auxiliary predicates \citep{DBLP:journals/tplp/CalimeriFGIKKLM20}.

We use the hybrid 
system \clingoDL~\citep{janhunen2017clingo}\footnote{\url{https://github.com/potassco/clingo-dl}.} that extends 
the ASP solver \clingo 
by difference logic to deal with timing constraints.
A difference constraint is an expression of the form $u - v \leq d$, where $u$ and $v$ are integer variables and $d$ is an integer constant. 
In contrast to unrestricted integer constraints, systems of difference constraints are solvable in polynomial time. 
The latter are expressed in \clingoDL using \emph{theory atoms}~\citep{gebser2016theory}. 
That job \verb!j! starts after its release time, say 10, can be expressed as
\verb!&diff{ 0 - start(j) } <= -10!.
Here, \verb!0! and \verb!start(j)! are integer variables, where \verb!0! has a fixed value of $0$; thus \verb!start(j)! 
must  
be 
at least $10$.

We will also use an extension for \emph{heuristic-driven solving}~\citep{gebser2013domain} that allows
to incorporate domain heuristics into ASP. These heuristics do not change the answer sets of a program but modify internal solver heuristics to bias search.
The general form of a heuristic directive is 
\begin{equation*}
    %\small
    \verb!#heuristic ! A \verb! : ! B\verb!. ! \verb![!w\verb!@!p\verb!,!m\verb!]!
\end{equation*}
where $A$ is an atom, $B$ is a rule body, and $w$, $p$, $m$ are terms.
In particular, $w$ is a weight, $\verb!@!p$ is an optional priority, and $m$ is a modifier
like \verb!true! or \verb!false!. We will 
provide further details when we introduce specific heuristic rules later on.

Further features of the input language,  like 
aggregation 
and optimisation statements, will be
explained as we go.

%%%%%%%%%%%%%%%%%%%%%%%%%%%%%%%%%%%%%%%%%%%%%%%%%%%%%%%%%%%%%%%%%%%%%%%%%%%%%%%%%%%%%
\section{Problem Statement}\label{sec:problem}
%%%%%%%%%%%%%%%%%%%%%%%%%%%%%%%%%%%%%%%%%%%%%%%%%%%%%%%%%%%%%%%%%%%%%%%%%%%%%%%%%%%%%

\begin{figure}
    \centering
    \begin{subfigure}{0.33\textwidth}\centering
    \includegraphics[scale=0.12]{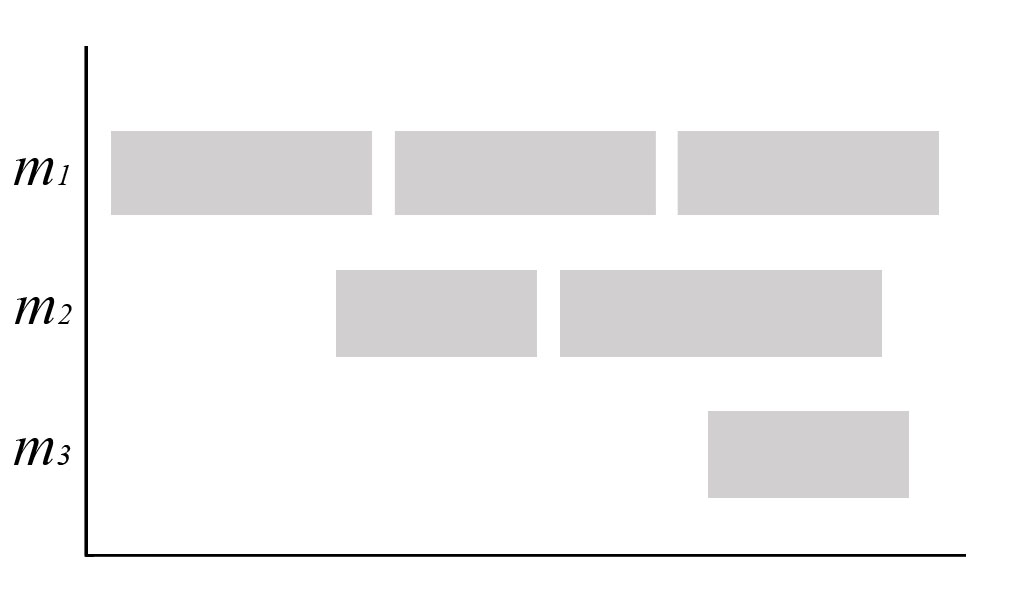}
    \caption{}
    \end{subfigure}\centering
    \begin{subfigure}{0.33\textwidth}\centering
    \includegraphics[scale=0.12]{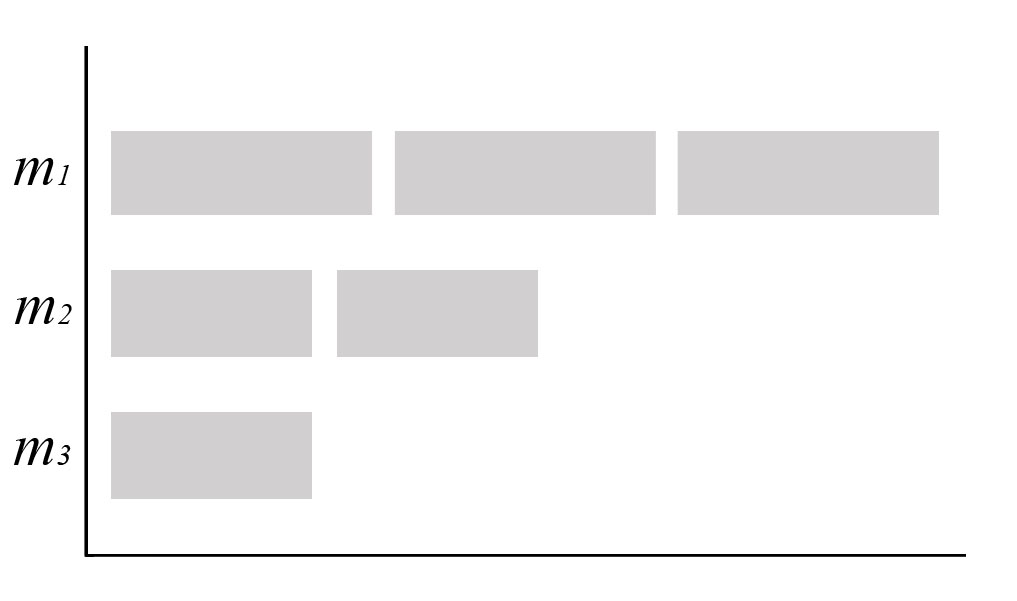}
    \caption{}
    \end{subfigure}
    \begin{subfigure}{0.33\textwidth}\centering
    \includegraphics[scale=0.12]{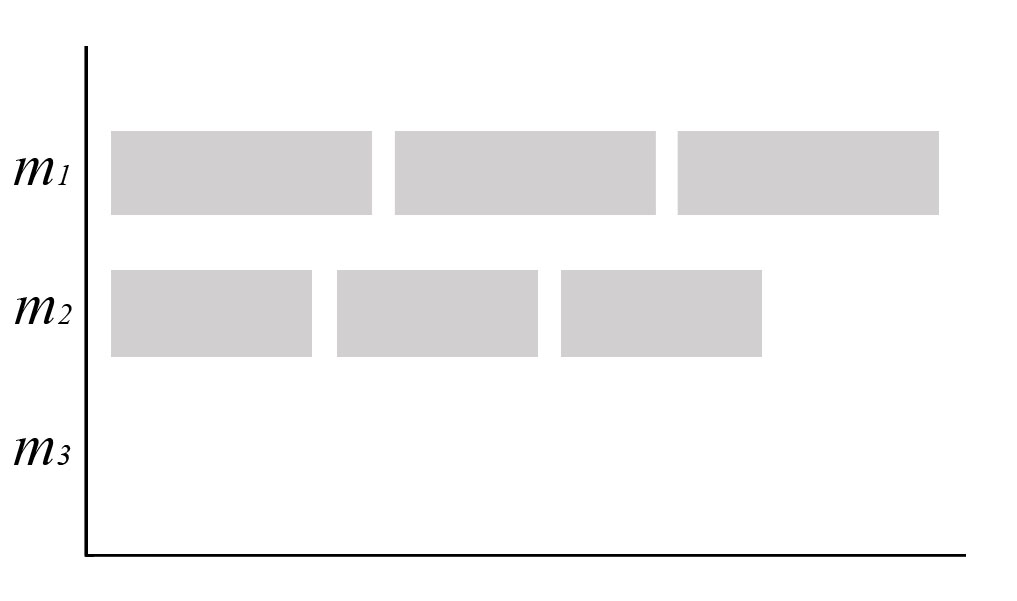}
    \caption{}
    \end{subfigure}
    \caption{Different schedules involving three machines and six jobs.}\label{fig:schedules}
\end{figure}

We study the following scheduling problem.
Given $m$ machines and $n$ jobs, 
every job needs to be processed by a single machine, and every machine can process at
most one job at a time; preemption is not allowed.
Some machines can only handle certain jobs, such that 
from the view of the latter, $\mcap{j}$
is the set of machines that can process job $j$.

We assume that a \emph{release date} $r_{j,k}$ is specified for every job $j$ and machine $k$ as a non-negative integer. 
Release dates are machine dependent because 
transportation time for jobs to the machines depends on the transport system and their location.
No job can start before its release date.

A specified amount of time may be required to change from one job to the next one. Specifically, we assume that 
$s_{i,j,k}$ is the time needed to set up job $j$ directly after job $i$ on machine $k$. Consequently, these times are referred to as \emph{sequence-dependent setup times}.  
Every job $j$ has a positive \emph{duration} $d_{j,k}$ that depends on the machine $k$ it is assigned to.

A \emph{schedule} $S$ for a problem instance is defined by:
\begin{enumerate}
\item an assignment $a$ that maps each job $j$ to a
machine $k \in \mcap{j}$ capable of processing it; 
\item for each machine $k$, a total order $\preceq_k$ on the set $J$ of jobs assigned to the machine via $a$. 
Relation $\preceq_k$ determines the sequence in which the jobs in $J$ are processed on $k$.
\end{enumerate}

If each job can be processed by some machine, then some schedule for the problem instance exists. 

Assume that $j_1, \ldots, j_l$ 
is the processing sequence of the jobs assigned to machine $k$ in a given
schedule. 
The \emph{processing time} $p_{j_i}$ of a job $j_i$ is its duration plus the setup time for its predecessor (if one exists); i.e.,
$p_{j_1} = d_{{j_1},k}$ and $p_{j_i} = s_{j_{i-1},j_i,k} + d_{j_i,k}$, for $i > 1$.
The \emph{start time} $\mathit{st}_{j_i}$ of job $j_i$ is $r_{j_i,k}$ if $i=1$, and $\mathit{max} (r_{j_i,k},\mathit{st}_{j_{i-1}} + p_{j_{i-1}})$ for $i > 1$. 
The \emph{completion time} $c_{j_i}$ of job $j_i$ is $st_{j_i}+p_{j_i}$.

The \emph{machine span} of $k$, $\mspan{k}$, is the completion time of the last job $j_l$ on $k$.
A common optimisation criterion is to search for a schedule with a small \emph{makespan}, which is 
the largest machine span of the schedule.

Versions of this problem have been extensively studied in the literature~\citep{allahverdi2015third}. 
The problem presented here abstracts the actual problem statement 
at Bosch
to its most essential elements.
We left out some details due to confidentiality. Other elements, like due dates or manual labor costs
have been omitted as they are not relevant for the objective function studied in this paper.

%%%%%%%%%%%%%%%%%%%%%%%%%%%%%%%%%%%%%%%%%%%%%%%%%%%%%%%%%%%%%%%%%%%%%%%%%%%%%%%%%%%%%
\subsection{The Lexicographical Makespan Objective}\label{sec:objective}
%%%%%%%%%%%%%%%%%%%%%%%%%%%%%%%%%%%%%%%%%%%%%%%%%%%%%%%%%%%%%%%%%%%%%%%%%%%%%%%%%%%%%

Recall that we are interested in computing schedules that maximise the throughput, but 
high machine dedication and rescheduling due to sudden machine failure 
render minimal makespan as a single objective function suboptimal. 

Figure~\ref{fig:schedules} (a) illustrates this: assume all jobs scheduled to machine $m_1$ cannot be processed by any other machine. Thus, $m_1$ will always determine the makespan and the remaining jobs can be put almost arbitrarily on the remaining machines. This can lead to unnecessary workload on 
these machines. 
A more severe problem 
is when machines suddenly fail 
and their jobs need to be rescheduled.
To cope with events like machine failure, the domain experts formulated the requirement that
``all machines should complete as early as possible''
with the intention to give the scheduler maximal freedom in rearranging jobs
with minimal decrease in throughput.

We next
define the \emph{lexicographical makespan} for lexicographical optimisation of machine spans to obtain robust schedules~\citep{letsios2021exact}.

\begin{definition}\label{def:lexmakespan}
Given a schedule $S$ involving $m$ machines, the \emph{lexicographical makespan}, or lex-makespan for short,
of $S$ is the tuple $ms(S) = (c_1,\ldots,c_m)$ of all the machine spans of $S$ in non-ascending order.
\end{definition}

In this definition, $c_1$ is a maximal machine span and hence corresponds to the makespan.

For schedules $S$ and $S'$ involving $m$ machines each, $S$ has a smaller lex-makespan than $S'$ if 
$ms(S)$ is smaller than
$ms(S')$ under lexicographical order, i.e., on the least index $i$ where $ms(S)=(c_1,\ldots,c_m)$ and $ms(S')=(c'_1,\ldots,c'_m)$ disagree, we have $c_i<c'_i$. 
For a set $\mathcal{S}$ of schedules, $S \in \mathcal{S}$ is then optimal if 
$ms(S)$ is minimal over all schedules in $\mathcal S$.

Consider Figure~\ref{fig:schedules} for illustration. We would prefer schedule (b)  over (c) under the lex-makespan objective. For both schedules, the lex-makespan is given by the machine spans of $m_1$, $m_2$, and $m_3$ in that order. 
Both schedules have the same makespan,
but schedule (b) has a smaller machine span for $m_2$. If machine $m_1$ fails and
most of the jobs can only be rescheduled to machine
$m_2$, schedule (b) would indeed be advantageous. It happens also earlier for schedule (b) that machines $m_2$ and $m_3$ complete all their jobs and are therefore free if new jobs need to be scheduled. 

To describe the dynamics of a schedule, we define, 
for a time point $t$ and schedule $S$, $M(S,t)$ as the number of machines that complete at or before $t$. We then obtain:
\begin{proposition}\label{prop:lexmakespan}
Let $S$ and $S'$ be two schedules for some problem instance.
Then, 
$ms(S) < ms(S')$ iff there is 
a time point $t$ such that 
$M(S,t) > M(S',t)$, and, 
for every $t' > t$, $M(S,t') \geq M(S',t')$.
\end{proposition}
\begin{proof}
Let $s = (c_1,\ldots,c_m)$ be the lexical makespan of $S$ and
$s' = (d_1,\ldots,d_m)$ the one of $S'$.
Let ($\ast$) stand for the right-hand side of the proposition:
there is a time point $t$ such that
$M(S,t) > M(S',t)$ and for any $t' > t$, $M(S,t') \geq M(S',t')$.

($\Rightarrow$)\quad Assume that $s < s'$.
Consider the least $i \in \{ 1, \ldots, m\}$ with $c_i < d_i$.
Observe that $M(S,c_i) \geq m-i+1$ while $M(S',c_i) = m-i$.
Thus, there is a time point $t=c_i$ such that $M(S,t) > M(S',t)$.
Since $c_j = d_j$ for any $j < i$, it follows that,
for any time point  $t' > t$, $M(S,t') \geq M(S',t')$. Hence, ($\ast$) holds.

($\Leftarrow$)\quad Towards a contradiction, assume that both $s \geq s'$ and
($\ast$) hold.
We distinguish between the two cases $s = s'$ and $s' < s$.
If $s = s'$ then there cannot be any time point $t$ with $M(S,t) >  M(S',t)$, a contradiction to ($\ast$).
If $s' < s$, the left-to-right side of the proposition implies that
there is a time point $t$ such that $M(S',t) >  M(S,t)$, and,
for any $t' > t$, $M(S',t') \geq M(S,t')$. Again, a contradiction to ($\ast$).
\end{proof}

For problems involving many machines, hierarchically minimising all the machine spans can be excessive if the overall makespan is dominated by few machines only.
However, comparing lex-makespans allows for  a rather natural parametrisation, namely an integer $l$ that defines the number of components to consider in the comparison. 

\begin{definition}\label{def:paramcomp}
Given schedules $S$ and $S'$ involving $m$ machines each and 
an integer $l$, $1 \leq l \leq m$, we say 
$ms(S) = (c_1,$ \ldots, $c_m)$ is smaller than
$ms(S') = (c'_1,\ldots,c'_m)$ under \emph{parametrised lexicographical order},
 in symbols $ms(S) \leq_l ms(S')$, if under lexicographical order
$(c_1,\ldots, c_l) \leq (c'_1,\ldots, c'_l)$. 
\end{definition}

Note that for a schedule with $m$ machines, we obtain the makespan objective if $l=1$ and the full lex-makespan if $l=m$.

%%%%%%%%%%%%%%%%%%%%%%%%%%%%%%%%%%%%%%%%%%%%%%%%%%%%%%%%%%%%%%%%%%%%%%%%%%%
\section{An Exact ASP Model with Difference Logic}\label{sec:exact}
%%%%%%%%%%%%%%%%%%%%%%%%%%%%%%%%%%%%%%%%%%%%%%%%%%%%%%%%%%%%%%%%%%%%%%%%%%%

A problem instance is described by ASP facts using some fixed predicate 
names. We illustrate this by an example with
one machine $m_1$ and two jobs $j_1$, $j_2$.
The machine is capable of processing all jobs and all release dates are $0$.
The setup time is $4$ when changing from job $j_1$ to $j_2$ and
$2$ 
vice versa. Both jobs have 
duration 5. 
The  according 
facts are 

{ %\small
\begin{Verbatim}
machine(m1). 
cap(m1,j1). cap(m1,j2).
job(j1). duration(j1,m1,5). release(j1,m1,0).
job(j2). duration(j2,m1,5). release(j2,m1,0).
setup(j1,j2,m1,4). setup(j2,j1,m1,2).
\end{Verbatim}
}
\noindent
Any problem instance can be described using this format.

%%%%%%%%%%%%%%%%%%%%%%%%%%%%%%%%%%%%%%%%%%%%%%%%%%%%%%%%%%%%%%%%%%%%%%%%%%%
\begin{figure}[t]\centering
\begin{minipage}{0.9\textwidth}
{\small
\begin{Verbatim}[numbers=left]
1 {asg(J,M): cap(M,J)} 1 :- job(J).

before(J1,J2,M) | before(J2,J1,M) :- asg(J1,M), asg(J2,M), J1 < J2.

1 {first(J,M): asg(J,M)} 1 :- asg(_,M).
1 {last(J,M): asg(J,M)} 1 :- asg(_,M).
1 {next(J1,J2,M): before(J1,J2,M)} 1 :- asg(J2,M), not first(J2,M).
1 {next(J2,J1,M): before(J2,J1,M)} 1 :- asg(J2,M), not  last(J2,M).
:- first(J1,M), before(J2,J1,M).
:- last(J1,M), before(J1,J2,M). 
                  
&diff{0 - c(J1)} <= -(T+D+S) :- asg(J1,M), next(J3,J1,M), 
                                setup(J3,J1,M,S), duration(J1,M,D), 
                                release(J1,M,T).  
&diff{c(J2) - c(J1)} <= -(P+S) :- before(J2,J1,M),  next(J3,J1,M), 
                                  setup(J3,J1,M,S), 
                                  duration(J1,M,P).
&diff{0  - c(J1)} <= -(T+D) :- asg(J1,M), duration(J1,M,D), 
                               release(J1,M,T). 
&diff{c(J)  - cmax} <= 0 :- job(J).

&diff{c(J2) - c(J1)} <= -P :- before(J2,J1,M), duration(J1,M,P).

1 {span(M,T): int(T)} 1 :- machine(M).
&diff{c(J) - 0} <= S :- asg(J,M), span(M,S).
#minimize{T@T,M: span(M,T)}.
\end{Verbatim}
}
\end{minipage}

\caption{ASP encoding with difference logic for lex-makespan optimisation.}\label{fig:clingoDL}
\end{figure}
%%%%%%%%%%%%%%%%%%%%%%%%%%%%%%%%%%%%%%%%%%%%%%%%%%%%%%%%%%%%%%%%%%%%%%%%%%%%

Next, we present the ASP encoding  for computing minimal schedules.
The entire program is given in Figure~\ref{fig:clingoDL}.

The encoding consists of three parts: Lines 1--10 qualitatively  model  feasible sequences of jobs on machines, while the quantitative model for completion times is realised in Lines 12-22 with difference logic; we avoid by this 
doing integer arithmetic in the Boolean ASP constraints, which 
would blow up the size of the grounding.
Finally, the optimisation is accomplished by Lines 24-26. Intuitively, we guess an upper bound for each machine span, which is then minimised. The directive in Line 26 assigns each span a priority equal to their value, thus ensuring that the highest span is the most important followed by the second highest and so forth.

The first line of Figure~\ref{fig:clingoDL} expresses
that each job is assigned to a machine capable of processing it.
The notation
{\tt asg(J,M)\,{:}\,cap(M,J)}
means that in the grounding step 
for each value {\tt j} of the global variable {\tt J} (as it occurs in the body),  
\verb!asg(J,M)! is replaced
by 
all atoms
{\tt asg(j,m)} for which \verb!cap(j,m)! can be derived.

We further require that the jobs assigned to a machine are totally ordered. 
That is, for any two distinct such jobs $j_1$ and $j_2$,
either $j_1 \prec j_2$ or $j_2 \prec j_1$ holds.
This is achieved by the rule in Line 3.
In Lines 5--6, 
the predicates \verb!first/2! and \verb!last/2!, representing the first and last job on each machine, respectively, are defined.
Constraints in Lines 9-10 ensure that this selection is compatible with the order given by \verb!before/3!.
Furthermore, each job except the last (resp.\ first) has a unique successor (resp.\ predecessor);
this is captured by
\verb!next/3! in Lines 7-8. 

We use difference logic to express  that jobs are put on the machines in the order defined by $\verb!next/3!$.
The rules in Lines 12--19 closely follow respective definitions from  Section~\ref{sec:problem}.
Line 20 defines \verb!cmax! as an upper bound of any completion time. In any answer-set, \verb!cmax! will be the actual makespan since the solver will always instantiate integer variables with the smallest value possible.
The redundant rule in Line 22 helps the solver to further prune the search space.

For optimisation, we ``guess'' a span for each machine in Line~24.
Here \verb!int/1! is assumed to 
provide a bounded range of integers.
In Line 25, we enforce that machines complete not later than the guessed spans.
The actual objective function is defined by the last line of Figure~\ref{fig:clingoDL} notably concise: any machine contributes its span $c$ to a cost function at priority level $c$. 
The cost function accumulates contributing values and the solver minimises answer sets by lexicographically comparing cost tuples 
ordered by priority. 

The formal correctness of the encoding is assured by the following proposition.

\begin{proposition}\label{prop:asp-encoding}
For every problem instance $I$, 
the schedules of $I$ with minimal lex-makespan are in one-to-one correspondence with the optimal answer sets of the program $P_I$ consisting of the rules in Figure~\ref{fig:clingoDL} augmented with
the fact representation of $I$.
\end{proposition}

\begin{proof}[Proof (Sketch)]
Given a (correct) schedule $S$ for a problem instance $I$, we can define an interpretation $M_I$ that is an answer set of the program $P_I$. 
To this end, we include in $M_I$ all atoms \verb|asg(j,k)| such that job $j$ is assigned to machine $k$ in $S$; all atoms \verb|before(j1,j2,k)| such that jobs $j1$ and $j2$ are assigned in $S$ to the same machine $k$ and $j1$ starts before $j2$; all atoms \verb|next(j1,j2,k)| such that jobs $j1$ and $j2$ are assigned in $S$ to the same machine $k$ and $j1$ starts immediately before $j2$; the atoms \verb|first(j,k)| and \verb|last(j,k)| where $j$ is in $S$ the first respectively last job on machine $k$; and the atom \verb|span(k,t)| for each machine $k$, where $t$ is $span(k)$ according to $S$. Furthermore, each integer variable 
\verb|c(j)| is set to the completion time of job $j$ in $S$;  \verb|cmax| equals the maximal completion time. It can then be shown that $M_I$ is an answer set of $P$,  
according to the semantics of \clingoDL. In turn, it can be shown that every answer set $M$ of the program $P_I$ encodes a correct schedule $S$ for the problem instance $I$, which is given by the atoms \verb|asg(j,k)|, \verb|before(j1,j2,k)| in $M$. 

 Thus, the correct schedules for $I$ correspond to the answer sets of $P_I$. By the statement \verb|#minimize{T@T,M: span(M,T)}|, those answer sets $M$ are optimal where 
 the vector $v_M = (span(k_1), span(k_2),\cdots, span(k_m))$ of all machine spans in the encoded schedule in decreasing order, i.e., $span(k_{j})\geq span(k_{j-1})$ for all $m \geq j > 1$, is lexicographical minimal; indeed, if vector $v_M$ is smaller than vector $v_{M'}$, then the value of the objective function for $M$ is smaller than the one for $M'$. Consequently, the optimal answer sets $M$ of $P_I$ correspond one-to one to the 
 schedules for $I$ with minimal lex-makespan.
\end{proof}

%%%%%%%%%%%%%%%%%%%%%%%%%%%%%%%%%%%%%%%%%%%%%%%%%%%%%%%%%%%%%%%%%%%%%%%%%%%%%
\subsection{Direct Multi-shot Optimisation}

The performance bottleneck for the ASP approach from the previous section is grounding.
In particular, the definition of the machine spans 
must be 
grounded over the entire relevant
integer
range. 
We can define machine spans in difference logic as bounds on completion times  similar to the makespan:
{%\small
\begin{verbatim}
&diff{   0 - span(M)} <= 0 :- machine(M).
&diff{c(J) - span(M)} <= 0 :- asg(J,M).    
\end{verbatim}
}
However, multi-objective optimisation for integer variables with priorities is unfortunately not supported in the current version (1.1.1) of \clingoDL.
Schedules with minimal lex-makespan are still computable using multiple 
solver calls and incrementally adding constraints.

%%%%%%%%%%%%%%%%%%%%%%%%%%%%%%%%%%%%%%%%%%%%%%%%%%%%%%%%%%%%%%%
\SetKwInput{KwInput}{Input}   % Set the Input
\SetKwInput{KwOutput}{Output} % set the Output

\begin{algorithm}[t]
\DontPrintSemicolon
\KwInput{model $M$ involving $m$ machines and
         parameter $l$ with $1 \leq l \leq m$
}
\KwOutput{schedule $R$ for $M$ with parametrised lex-makespan $(c_1,\ldots,c_l)$\\
\BlankLine
$\solve{M} \ldots $ returns a solution for $M$ or $\emptyset$ if none is found within fixed resource limits \\
\BlankLine
$\bound{i \circ b}$,  $\circ \in \{<,\leq\} \ldots $ constraints enforcing that ${c_i} \circ b$ for the 
lex-makespan  $(c_1,\ldots,c_m)$
\BlankLine
\\ 
\BlankLine
}

$(c_1,\ldots,c_l) \leftarrow (0,\ldots,0)$\;
$R \leftarrow \solve{M}$\;

\For{$i \leftarrow 1$ \KwTo $l$} {
    $S \leftarrow R$\;
    \While{$S \neq \emptyset$}{
             $R \leftarrow S$\; 
             $c_i \leftarrow\ $ $ith$ element of the lex-makespan of $S$\; 
             $S \leftarrow \solve{M \cup \bound{i < c_i} }$\;
    }
    
    $M \leftarrow M \cup \bound{i \leq c_i}$\;
}
\Return{$R$} with lex-makespan $(c_1,\ldots,c_l)$\;

\caption{Lex-Makespan Optimisation}\label{alg:exact}
\end{algorithm}
%%%%%%%%%%%%%%%%%%%%%%%%%%%%%%%%%%%%%%%%%%%%%%%%%%%%%%%%%%%%%%%

To this end, we present Alg.~\ref{alg:exact} for lex-makespan minimisation by using multiple solver calls. 

Alg.~\ref{alg:exact} is a  standard way for multi-objective minimisation by 
doing a (highest priority first) hierarchical descent. 
Due to symmetries, showing a lack of solutions is usually more costly for this problem than finding
one; this makes alternative strategies
with fewer expected solver calls like binary search or exponentially increasing search steps less attractive. 

We use \clingoDL and the encoding from Figure~\ref{fig:clingoDL} without the optimisation statement in Lines 21--23
to implement $\solve{M}$ in Alg.~\ref{alg:exact}.
A handy feature is that \clingoDL supports
\emph{multi-shot solving}~\citep{gebser2019multi} where parts of the solver state are kept throughout multiple runs, thereby saving computational resources.
Notably $\solve{M}$ in Alg.~\ref{alg:exact} does not limit us to use ASP solver. We can in principle use any exact method that is capable of producing solutions for a model $M$ in the input language of the respective system.

The constraints for $\bound{i \leq b}$ are quite easy to express in ASP:
that the $i$-th component of the lex-makespan is smaller than or equal to $b$ is equivalent to 
enforcing that at least $m-i+1$ machines have a span 
of at most $b$.
We can encode the latter 
by non-deterministically selecting  $m-i+1$ machines and enforcing
that  
they complete not later than $b$:

{%\small 
\begin{Verbatim}
(m-i+1) {sel(M): machine(M)}.
&diff{span(M) - 0} <= b :- sel(M).
\end{Verbatim}
}

While Alg.~\ref{alg:exact} is guaranteed to return a schedule with minimal lex-makespan when resources for $\solve{M}$ are not limited, we will in practice restrict the time spent for search in $\solve{M}$ by a suitable time limit.

%%%%%%%%%%%%%%%%%%%%%%%%%%%%%%%%%%%%%%%%%%%%%%%%%%%%%%%%%%%%%%%%%%%%%%%%%%%
\subsection{Domain-specific Heuristics}

We use two domain heuristics to improve performance by guiding search more directly to promising areas of the search space.
Both heuristic directives 
use the modifier \verb!true!: Whenever an atom needs to be assigned a truth value,
the solver will pick the one with the highest weight among the ones with highest priority and assigns it to true at first.

Recall that job durations depend on the machines. 
The first heuristic expresses the idea to assign jobs to machines if their duration is low on that machine. 
{%\small
\begin{verbatim}
#heuristic asg(J,M): duration(J,M,D), 
                     maxDuration(J,F), W=F-D. [W@2,true]
maxDuration(J,M) :- job(J), M = #max{D: duration(J,_,D)}.
\end{verbatim}
}
Here, \verb!maxDuration/2! defines the longest duration of a given job over all machines.
Then, the heuristic directive gives a high weight to an atom \verb$asg(j,m)$ if the 
duration of $j$ is low relative to its maximal duration. The priority level of this rule is $2$; this means that the solver will try to assign jobs to machines before deciding on other atoms.

The second heuristic directive affects how jobs are put on machines.
We want to avoid large setup times and follow an analogous strategy as for the first heuristic:

{%\small
\begin{verbatim}
#heuristic next(J,K,M): setup(J,K,M,S), maxSetup(K,M,T), 
                        cap(J,M), cap(K,M), W=T-S. [W@1,true]
maxSetup(J,M,S) :- job(J), machine(M), 
                   S = #max{T: setup(_,J,M,T)}.
\end{verbatim}
}
Atom \verb!next(j,k,m)! gets a high weight if putting job $j$ before $k$ results in a relatively small setup time. We also need to make sure that machine $m$ is actually capable of processing  both jobs. The order of the jobs has a lower priority than the machine assignment.

\subsection{Plain Clingo Model}\label{sec:plainasp}

%%%%%%%%%%%%%%%%%%%%%%%%%%%%%%%%%%%%%%%%%%%%%%%%%%%%%%%%%%%%%%%%%%%%%%%%%%%
\begin{figure*}\centering
\begin{minipage}{.9\textwidth}
{\small
\begin{Verbatim}[numbers=left]
1 {asg(J,M) : cap(M,J)} 1 :- job(J).
1 {first(J1,M) : asg(J1,M)} 1 :- #count{J2 : asg(J2,M)} > 0.
1 {last(J1,M) : asg(J1,M)} 1 :- #count{J2 : asg(J2,M)} > 0.
1 {next(J1,J2,M) : asg(J1,M)} 1 :- asg(J2,M), not first(J2,M).
1 {next(J1,J2,M) : asg(J2,M)} 1 :- asg(J1,M), not  last(J2,M).
reach(J,M)  :- first(J,M).
reach(J2,M) :- reach(J1,M), next(J1,J2,M).
:- asg(J,M), not reach(J,M).

process(J,T)    :- first(J,M),    duration( J,M,T).
process(J2,D+S) :- next(J1,J2,M), duration(J2,M,D), 
                   setup(J1,J2,M,S).
start(J,T)  :- int(T), first(J,M), release(J,M,T).
start(J2,T) :- int(T), next(J1,J2,M), 
               T = #max{R : release(J2,M,R) ; C : compl(J1,C)}.
compl(J,S+P) :- int(T), start(J,S), process(J,P).
span(M,T)    :- int(T), last(J,M), compl(J,T).

#minimize{ T@T,M : span(M,T) }.
\end{Verbatim}
}
\end{minipage}
\caption{Plain ASP encoding for lexical makespan minimisation.}\label{fig:asp-schedules}
\end{figure*}
%%%%%%%%%%%%%%%%%%%%%%%%%%%%%%%%%%%%%%%%%%%%%%%%%%%%%%%%%%%%%%%%%%%%%%%%%%%%

\noindent
The ASP encoding in Figure~\ref{fig:clingoDL} can also be formulated without difference logic which makes it suitable as input for the ASP solver \clingo.
This ASP program for computing minimal schedules is given in Figure~\ref{fig:asp-schedules}.
There, completion times are computed directly with arithmetic aggregates. While this approach works for small instances, it does not scale for larger integer domains.

%%%%%%%%%%%%%%%%%%%%%%%%%%%%%%%%%%%%%%%%%%%%%%%%%%%%%%%%%%%%%%%%%%%%%%%%%%%
\section{ASP-based Approximation}\label{sec:approx}
%%%%%%%%%%%%%%%%%%%%%%%%%%%%%%%%%%%%%%%%%%%%%%%%%%%%%%%%%%%%%%%%%%%%%%%%%%%

While the exact methods from the previous section have the advantage that we can run a solver until we find a guaranteed optimal solution, this works only for very small problem instances. Finding good solutions within a time limit is in practice more important than showing optimality.
This is what the ASP-based approximation method we discuss next are designed to accomplish.

There is a simple way to turn the exact encoding from 
Figure~\ref{fig:clingoDL} into an approximation that scales better to larger instances. It has been introduced 
for \clingoDL optimisation in the context of train scheduling~\citep{abels2019train}, and we apply it for our machine scheduling application.
Recall that we use \verb!int/1! to define a range $[0, 1, \ldots, n]$ of integers from which potential bounds for individual machine spans are taken from. 
We can instead consider integers from $[0, 1 \cdot g, \ldots, i \cdot g]$ where $g$ is the granularity of the approximation, and $i$ is chosen such that $n \leq i \cdot g$; $i$ can be significantly smaller than $n$, depending on
$g$, and thus reduce the search space and the size of the grounding. A larger $g$ makes the approximation more coarse, a smaller $g$ makes it closer to the exact encoding.  
We will compare the exact encoding and this approximation in Section~\ref{sec:experiments}.

%%%%%%%%%%%%%%%%%%%%%%%%%%%%%%%%%%%%%%%%%%%%%%%%%%%%%%%%%%%
\subsection{Multi-shot Approximation}

We next present a variant of Alg.~\ref{alg:exact} for approximation, where we assume that we have an exact optimiser $\opt{\cdot}$ which is good at finding schedules with small makespans.  This optimiser can then be employed to compute schedules with small lex-makespans. The specifics are presented as Alg.~\ref{alg:approx}

%%%%%%%%%%%%%%%%%%%%%%%%%%%%%%%%%%%%%%%%%%%%%%%%%%%%%%%%%%%%%%%
\begin{algorithm}[t]
\DontPrintSemicolon
\KwInput{model $M$ involving $m$ machines and parameter $l$ with $1 \leq l \leq m$ }
\KwOutput{schedule $S$ for $M$ with parametrised lex-makespan $(c_1,\ldots,c_l)$\\
\BlankLine
$\opt{M} \ldots $ returns the best solution for $M$ found within fixed resource limits \\
\BlankLine
}

$(c_1,\ldots,c_l) \leftarrow (0,\ldots,0)$\;
$S \leftarrow \emptyset$\;

\For{$i \leftarrow 1$ \KwTo $l$} {

   $S \leftarrow \opt{M}$\;
   
   $c_i \leftarrow\ $ makespan of $S$\; 
   remove some machine $k$ that completes at $c_i$ and all jobs assigned to $k$ from $M$\;
}

\Return{$S$}  

\caption{Lex-Makespan Approximation}\label{alg:approx}
\end{algorithm}
%%%%%%%%%%%%%%%%%%%%%%%%%%%%%%%%%%%%%%%%%%%%%%%%%%%%%%%%%%%%%%%

Algorithm~\ref{alg:approx}  uses 
$\opt{\cdot}$
to recompute and improve parts of a solution by
 fixing the jobs on the machine with highest span after each solver call.
Similar to Alg.~\ref{alg:exact}, we require multiple solver calls but 
with a profound difference: 
the number of solver calls to the  makespan optimiser is bounded by the number of machines, and
after each solver call, the problem instance is significantly simplified and thus easier to solve. 

As \clingoDL 
allows to directly 
minimise a single integer variable, 
we  can implement the makespan optimiser $\opt{\cdot}$ directly using our difference logic encoding and minimise \verb!cmax!.
However, we opted for using multi-shot solving again to reuse heuristic values and learned clauses from previous solver runs.

%%%%%%%%%%%%%%%%%%%%%%%%%%%%%%%%%%%%%%%%%%%%%%%%%%%%%%%%%%%%%%%%%%%%%%%%%%%%%%%%%%%%%
\section{Experimental Evaluation}\label{sec:experiments}
%%%%%%%%%%%%%%%%%%%%%%%%%%%%%%%%%%%%%%%%%%%%%%%%%%%%%%%%%%%%%%%%%%%%%%%%%%%%%%%%%%%%%

We 
now provide 
an experimental evaluation of the approaches for lex-makespan optimisation
from above. Notably, any approach that produces schedules with small makespan will
also produce 
small lex-makespans. 
Our primary goal is to investigate the difference in schedule quality when spending all resources for makespan optimisation versus distributing 
them for lex-makespan optimisation 
to different priorities. 
As we 
cannot disclose real instances from the semi-conductor production application, we
use random instances of realistic size and structure instead.
In addition to experiments with ASP-based solvers, to compare with other solving paradigms, 
we provide a solver-independent model in MiniZinc \citep{nethercote2007minizinc}. This model can be solved by many CP and MIP solvers. However, for a direct comparison, we also model our problem directly in \cplex and \cpoptimizer. 

%%%%%%%%%%%%%%%%%%%%%%%%%%%%%%%%%%%%%%%%%%%%%%%%%%%%
\subsection{Problem Instances}
\begin{table}%[h]
\centering\small

\tablefont{\begin{tabular}{@{\extracolsep{\fill}}rrrrrr}\toprule
      & M3    & M5     & M10     & M15     & M20 \\ \midrule
$m$   &  3    &  5     &  10     &  15     &  20 \\
$n$   & 5--50 & 10--50 & 50--200 & 100-200 & 150--200 \\ \bottomrule
\end{tabular}}
\caption{Machines ($m$) and jobs ($n$) per instance class.}
\label{tab:instance_classes}
\end{table}

We generated 500 benchmark instances of different sizes randomly. The random instance generator
is designed however to reflect relevant properties of the real instances.
The generator is based on previous work in the literature~\citep{vallada2011genetic}, but also produces instances with high machine dedication and amends older benchmarks, which were designed for different objective, with random release dates.
The 500 instances can be
grouped into five classes, shown in Table~\ref{tab:instance_classes},
of 100 instances each.
The instance generator as well as all the encoding and algorithms are online available.\footnote{
\url{http://www.kr.tuwien.ac.at/research/projects/bai/tplp22.zip}.
}

The machine capabilities 
were assigned uniformly at random for half of the instances in every class:
for each job, a random number of machines were assigned as capable. 
For the other half, we assigned the capabilities such that 80\% of the jobs can only be performed by 20\% of the machines. We refer to the latter setting as high-dedication and the former as low-dedication.

 For each job $j$ and any machine $k$, 
the  duration $d_{j,k}$, setup time $s_{j,i,k}$ for any other job $i$, and release date $r_{j,k}$ were drawn uniformly at random 
from $[10,500]$, $[0,100]$, and $[0,r_\mathit{max}]$, respectively, where 
$$r_\mathit{max}\!=\!\frac{1}{m}\sum_{1 \leq j \leq n}\!\frac{1}{|\mathit{cap}(j)|}\big(\!\sum_{k \in \mathit{cap}(j)}\!\!\! d_{j,k}+\hspace{-2ex}\sum_{\hspace{1.5ex}1 \leq j' \leq n, k \in \mathit{cap}(j')}\hspace{-4.5ex} s_{j',j,k}~\big).$$

%%%%%%%%%%%%%%%%%%%%%%%%%%%%%%%%%%%%%%%%%%%%%%%%%%%%%%%%%%%%%%%%%%%%%%%%%%%%%%%%%%
\subsection{A Solver-Independent MiniZinc Model}

As an alternative to ASP, we implemented a solver-independent model for schedule optimisation in the well-known high-level modelling language
MiniZinc for constraint satisfaction and optimisation problems.  
MiniZinc models, after being compiled into FlatZinc, can be used by a wide range of solvers. We provide a direct model that represents our best attempt at using MiniZinc.
It thus serves only as a first baseline and more advanced models may improve performance. 
Our model of the problem statement from 
Section~\ref{sec:problem} and the objective function  
are as follows.

For each job $i \in \{1, \dots , n\}$, we use the following decision variables:
    $a_i \in \{1, \dots, m\}$  for its assigned machine, 
    $p_i \in \{ 0, \dots , n\}$ representing its predecessor or 0 if it has none, and
    $c_i \in \{0, \dots, h\}$ denoting its completion time where $h$ is the scheduling horizon. 
For each machine $j \in \{ 1, \dots, m \}$, we use
    $s_j \in \{1, \dots, h\}$  denoting its span, and
    its level in the ordering of spans $l_j  \in \{ 1, \dots, m \}$. 

\begin{figure}
    \centering
    \begin{flalign}
&\mathtt{alldifferent\_except\_0}(p_{1 \leq i \leq n}) \label{mzn:predecessor_alldiff} \\
&\sum_{i=1}^n (p_i = 0) \leq \mathtt{nvalue}(a_{1 \leq i \leq m}) \label{mzn:num_first_jobs} \\
& a_i \in \mathit{cap}(i)  &  1 \leq i \leq n  \label{mzn:machine_cap} \\
& p_i \neq 0 \rightarrow  a_{i} = a_{p_i}  &  1 \leq i \leq n  \label{mzn:pred_same_machine} \\
& p_i \neq i  &  1 \leq i \leq n  \label{mzn:predecessor_diff} \\
& c_i \geq (\mathit{max}(c_j, \ r_{i,a_i}) + s_{j,i,a_i} + d_{i,a_i}) \cdot (p_i = j) & 1 \leq i,j \leq n, i \neq j \label{mzn:precedence} \\
& c_i \geq r_{i,a_i} + d_{i,a_i}  &  1 \leq i \leq n \label{mzn:release}\\
& s_k = \mathit{max}_{1 \leq i \leq n}(c_i \cdot (a_i = k)) & 1 \leq k \leq m \label{mzn:spans} \\
&\mathtt{alldifferent}(l_{1 \leq i \leq m}) \label{mzn:levels_alldiff}  \\
& l_{i} > l_{j} \rightarrow s_{i} \geq s_{j} & 1 \leq i,j \leq m \label{mzn:level_order} 
\end{flalign}
    \caption{Solver-independent MiniZinc model for lex-makespan optimisation.}
    \label{fig:model_minizinc}
\end{figure}

The constraints of the MiniZinc model are given in Figure~\ref{fig:model_minizinc}.
The global constraint (\ref{mzn:predecessor_alldiff}) ensures that no two jobs have the same predecessor, while 
(\ref{mzn:num_first_jobs}) enforces that the number of first jobs is less or equal to the number of assigned machines. The latter is determined through a global $\mathtt{nvalue}$ constraint returning the number of different values in $a_{1 \leq i \leq m}$. Constraint (\ref{mzn:machine_cap}) ensures that every job is assigned a capable machine, and constraint (\ref{mzn:pred_same_machine}) ensures that a job's predecessor is on the same machine. 
Constraint (\ref{mzn:predecessor_diff}) expresses that no job is its own predecessor, while
(\ref{mzn:precedence}) ensures that each job starts after its predecessor and 
the corresponding setup time.
Finally, 
(\ref{mzn:release}) enforces that every first job starts after its release date.

Defining an objective function for lex-makespan minimisation is more intricate. 
For this, we use constraints (\ref{mzn:spans}-\ref{mzn:level_order}). Here (\ref{mzn:spans}) defines the span for each machine to be the latest completion time of any job scheduled on it,
while (\ref{mzn:levels_alldiff}-\ref{mzn:level_order}) ensure that each machine is assigned a different level and the levels order the machines with respect to their spans.  

The objective function for minimising the lex-makespan can then be expressed as
$$\textstyle \mathit{min} \ \sum_{i=1}^m h^{l_i- 1} \cdot s_i.$$
Intuitively, the levels represent the priorities
for optimisation. By assigning the span of machine $i$ the weight $h^{l_i- 1}$, 
it is more important than all spans of  machines on lower levels.

Note that restricting the model to constraints (\ref{mzn:predecessor_alldiff}--\ref{mzn:release}) and using the objective $\mathit{min} \ \mathit{max}_{1\leq i \leq n}(c_i)$ expresses the scheduling problem with the standard makespan objective.

\subsection*{CPLEX and CP Optimizer Models}
We also encoded our problem in the modelling languages of \cplex and \cpoptimizer.
As for the MiniZinc model, the encodings represent our best attempts at using respective modelling languages.
The MIP model for \cplex uses the following variables.
\begin{itemize}
    \item $a_{j,i} \in \{0,1\}$ indicating that job $j$ is assigned to machine $i$,
    \item $p_{j,k,i} \in \{0,1\}$ representing that $k$ is the predecessor of $j$ ($k$ can be zero) on machine $i$,
    \item $c_j$ the completion time of job $j$,
    \item $s^M_i$ the span of machine $i$, 
    \item $s^L_l$ the span of level $l$, and 
    \item $l_{i,l} \in \{0,1\}$ indicating that machine $i$ has level $l$.
\end{itemize}
The last variable is obviously only needed for the lex-makespan objective.

\begin{figure}[t]
    \centering
\begin{flalign}
&\sum_{1 \leq i \leq m} a_{j,i} = 1 &  1 \leq j \leq n \label{cplex:assignment} \\ 
&\sum_{i \in \mathit{cap}(j)} a_{j,i} = 1 &  1 \leq j \leq n \label{cplex:capable} \\  
&\sum_{1 \leq j \leq n} p_{j,0,i} \leq 1 & 1 \leq i \leq m \label{cplex:first_jobs} \\
&\sum_{0 \leq k \leq n} p_{j,k,i} = a_{j,i} & 1 \leq j \leq n, \ 1 \leq i \leq m \label{cplex:pred} \\  
&\sum_{1 \leq k \leq n} p_{k,j,i} = a_{j,i} & 1 \leq j \leq n, \ 1 \leq i \leq m  \label{cplex:succ} \\
&c_j + h \cdot (1 - p_{j,k,i}) \geq c_k + d_{j,i} + s_{k,j,i} & 1 \leq j,k \leq n,\ 1 \leq i \leq m \label{cplex:start_after_pred} \\
& c_j \geq (r_{j,i} + d_{j,i}) \cdot a_{j,i}  & 1 \leq j \leq n,\ 1 \leq i \leq m \label{cplex:start_after_release1} \\
&c_j + h \cdot (1 - p_{j,k,i}) \geq r_{j,i} + d_{j,i} + s_{k,j,i} & 1 \leq j,k \leq n,\ 1 \leq i \leq m  \label{cplex:start_after_release2}\\
&s^M_i + h \cdot (1 - a_{j,i}) \geq c_j  & 1 \leq j \leq n,\ 1 \leq i \leq m \label{cplex:spans} \\
&(1 - l_{i,k}) \cdot h + s^L_k \geq s^M_i & 1 \leq k \leq l, \ 1 \leq i \leq m \label{cplex:level_spans}\\
& s^L_i \geq s^L_j & 1 \leq i < j \leq l \label{cplex:level_desc} \\
&\sum_{1 \leq k \leq l} l_{i,k} = 1 & 1 \leq i \leq m \label{cplex:machine_has_level} \\
&\sum_{1 \leq i \leq m} l_{i,k} = 1 & 1 \leq k \leq l \label{cplex:level_has_machine}
\end{flalign}
    \caption{MIP model used for lex-makespan optimisation in \cplex.}
    \label{fig:model_cplex}
\end{figure}

The model with all constraints is given in Figure~\ref{fig:model_cplex}. Constraints (\ref{cplex:assignment}) and (\ref{cplex:capable}) ensure that each job is assigned exactly one capable machine. Constraint~(\ref{cplex:first_jobs}) enforces that each machine has at most one first job. Constraint~(\ref{cplex:pred}) states that each job is either the first job on its assigned machine or has exactly one predecessor. The next constraint~(\ref{cplex:succ}) enforces that each job has at most one successor on its machine. Constraint~(\ref{cplex:start_after_pred}) ensures that each job starts after its predecessor. Job starting after their release is enforced through (\ref{cplex:start_after_release1}) and (\ref{cplex:start_after_release2}). Machine spans are ensured to be bigger than all completion times on the machine by Constraint~(\ref{cplex:spans}). Constraints~(\ref{cplex:level_spans}--\ref{cplex:level_has_machine}) make sure that the level spans correspond to the machine spans and are descending.

Optimising the lex-makespan objective can now be achieved by using the lexicographic minimisation functionality of \cplex to minimise the level spans $s^L_i (1 \leq i \leq l)$. 
We can also use this model to optimise makespan by simply dropping Constraints~(\ref{cplex:level_spans}-\ref{cplex:level_has_machine}) and minimising the maximum machine span.

The decision variables of the \cpoptimizer model are of a slightly different form.
\begin{itemize}
    \item $a_j$ indicates the assigned machine of job $j$ i.e. $1 \leq a_j \leq m$,
    \item $p_j$ is the predecessor of $j$ or zero if there is none,
    \item $c_j$ the completion time of job $j$,
    \item $s^M_i$ the span of machine $i$, 
    \item $s^L_l$ the span of level $l$, and 
    \item $l_i$ indicating the level of machine $i$.
\end{itemize}

The constraints for \cpoptimizer are given in Figure~\ref{fig:model_cpo}\footnote{For a proposition $P$, $[P] = 1$ if $P$ is true and $[P] = 0$ otherwise.}.
Constraint (\ref{cpo:pred_alldiff}) ensures that predecessors are all different or zero. The next constraint (\ref{cpo:pred_not_same}) enforces that no job is its own predecessor. Constraint (\ref{cplex:assignment}) makes sure that each job is assigned a capable machine. The number of first jobs is bounded by the number of assigned machines through Constraint (\ref{cplex:first_jobs}). Constraint (\ref{cpo:pred_same_machine}) ensures that the predecessor of a job is assigned the same machine. Jobs starting after their predecessor and their release is enforced through (\ref{cpo:start_after_pred}) and (\ref{cpo:start_after_release}). Machine spans are ensured to be bigger than all completion times on the machine by Constraint~(\ref{cpo:spans}). Lastly, Constraints (\ref{cpo:level_desc}--\ref{cpo:level_spans}) make sure that the level spans correspond to the machine spans and are descending.

Similarly to \cplex, optimising lex-makespan can be accomplished by using the lexicographic minimisation functionality of \cpoptimizer and the model can also be easily modified to minimise makespan.

\begin{figure}[ht]
    \centering
    \begin{flalign}
&p_j = 0 \lor p_k = 0 \lor p_j \neq p_k & 1 \leq j<k \leq n \label{cpo:pred_alldiff}\\
&p_j \neq j & 1 \leq j \leq n \label{cpo:pred_not_same}\\
&a_j \in \mathit{cap}(j) & 1 \leq j \leq n \label{cpo:assignment}\\
&\sum_{1 \leq i \leq m} [\mathtt{count}((a_j)_{1\leq j \leq n}, i) > 0] \geq \mathtt{count}((p_j)_{1\leq j \leq n}, 0) & \label{cpo:first_jobs}\\
&p_j \neq k \lor a_j = a_k & 1 \leq j,k \leq n,\ j \neq k \label{cpo:pred_same_machine}\\
&c_j \geq (\mathit{max}(r_{j,a_j}, c_k) + d_{j,a_j} + s_{k,j,a_j}) \cdot p_j = k & 1 \leq j,k \leq n, \ j \neq k \label{cpo:start_after_pred} \\
& c_j \geq r_{j,a_j} + d_{j,a_j} & 1 \leq j \leq n \label{cpo:start_after_release}\\
&s^M_i + (h \cdot [a_j \neq i]) \geq c_j & 1 \leq i \leq m, \ 1 \leq j \leq n \label{cpo:spans}\\
& s^L_i \geq s^L_j & 1 \leq i < j \leq l \label{cpo:level_desc} \\
&l_i \neq l_j & 1 \leq i < j \leq n \label{cpo:level_alldiff} \\
&[l_i \neq j] \cdot h + s^L_j \geq s^M_i & 1 \leq i \leq m, \ 1 \leq j \leq l \label{cpo:level_spans} 
\end{flalign}
    \caption{Native \cpoptimizer Model for lex-makespan optimisation.}
    \label{fig:model_cpo}
\end{figure}

%%%%%%%%%%%%%%%%%%%%%%%%%%%%%%%%%%%%%%%%%%%%%%%%%%%%%%%%%%%%%%%%%%%%%%%%%%%%%%%%%%
\subsection{Systems}

We use \clingoDL version (1.1.1) for $\solve{\cdot}$ and $\opt{\cdot}$ in Algs.~\ref{alg:exact} and \ref{alg:approx}, respectively.
For both algorithms, the time limit for optimising any level of the lex-makespan can be set to a geometric sequence with 
ratio $0.5$. 
Thus, half of the total time limit is spent on optimising the highest priority level,
a quarter on the next level etc.

We use a FlatZinc linearisation of the MiniZinc model to compare with four MIP/CP solvers (\cplex 12.10\footnote{\url{https://www.ibm.com/analytics/cplex-optimizer}.\label{fn:ibm}}, \cpoptimizer 20.1\footref{fn:ibm}, \gecode 6.3.0\footnote{\url{https://www.gecode.org}.} and  \ortools 7.8\footnote{\url{https://developers.google.com/optimisation}.}) against the hybrid ASP approach. All solvers could be used for makespan minimisation. Regarding the lex-makespan, \gecode could not produce any solutions due to numerical issues and both \cplex and \cpoptimizer wrongly reported optimality for some solutions. This is a technical issue that is probably due to the translation of MiniZinc model to FlatZinc or too high values for the lex-makespan objective function. Due to this, we also encoded the problem and both objectives in the native modelling languages of \cplex and \cpoptimizer.\footnote{The encodings are available online at \url{http://www.kr.tuwien.ac.at/research/projects/bai/tplp22.zip}.
} 
Those direct approaches had no problems with wrongly reported optimal solutions and their results are included below. 
In difference, \ortools had no issues with the MiniZinc model and was thus run using this model.
We generally used all solvers in single-threaded mode and with default settings.
          
%%%%%%%%%%%%%%%%%%%%%%%%%%%%%%%%%%%%%%%%%%%%%%
\subsection{Experimental Results and Discussion}

All experiments were conducted on a cluster with 13 nodes, where each node has two Intel Xeon CPUs E5-2650 v4 (max.~2.90GHz, 12 physical cores, no hyperthreading), and 256GB RAM. For each run, we 
set a memory limit of 20GB and all solvers only used one solving thread.
The 
application at the production site requires schedule computation within 300 seconds i.e. 5 minutes,
which we adopted as time limit. 
However, in order to offer additional insights, we also performed experiments with run times of up to 15 minutes.

%%%%%%%%%%%%%%%%%%%%%%%%%%%%%%%%%%%%%%%%%%%%%%%%%%%%%%%%%%%%%%%%%%%%%%%%%%%%%%%%%%%
\begin{figure}%
    \centering%
    \begin{subfigure}{\textwidth}\centering%
		\includegraphics[scale=0.4]{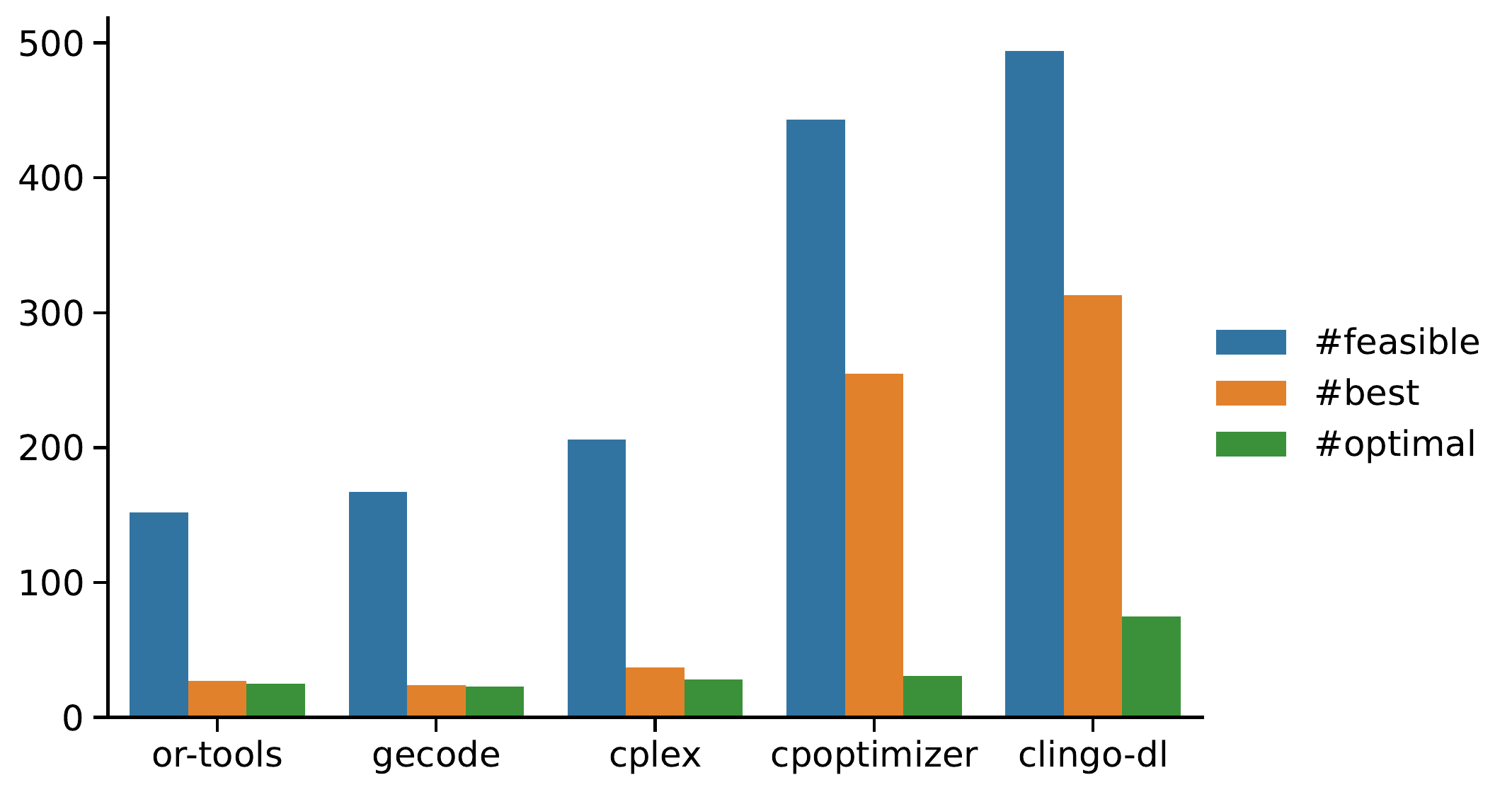}
    \end{subfigure}
    
    \centering
    \begin{subfigure}{\textwidth}\centering%
		\includegraphics[width=0.8\textwidth]{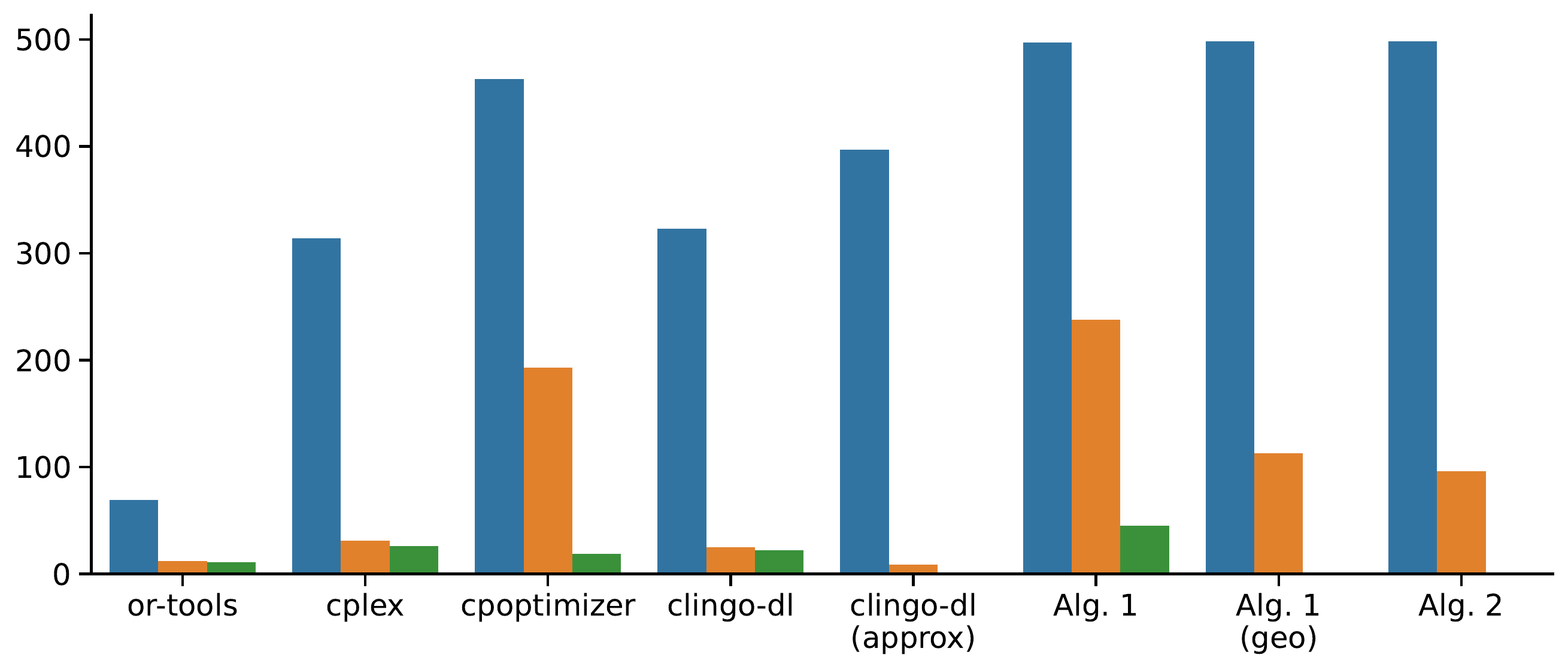}
    \end{subfigure}%
    \caption{Different systems on all instances for makespan (top) and lex-makespan (bottom) with a time limit of 5 minutes.}\label{fig:comparison-overall}%
\end{figure}
%%%%%%%%%%%%%%%%%%%%%%%%%%%%%%%%%%%%%%%%%%%%%%%%%%%%%%%%%%%%%%%%%%%%%%%%%%%%%%%%%%%

%%%%%%%%%%%%%%%%%%%%%%%%%%%%%%%%%%%%%%%%%%%%%%%%%%%%%%%%%%%%%%%%%%%%%%%%%%%%%%%%%%%
\begin{figure}%
    \centering%
    \begin{subfigure}{\textwidth}\centering%
		\includegraphics[scale=0.4]{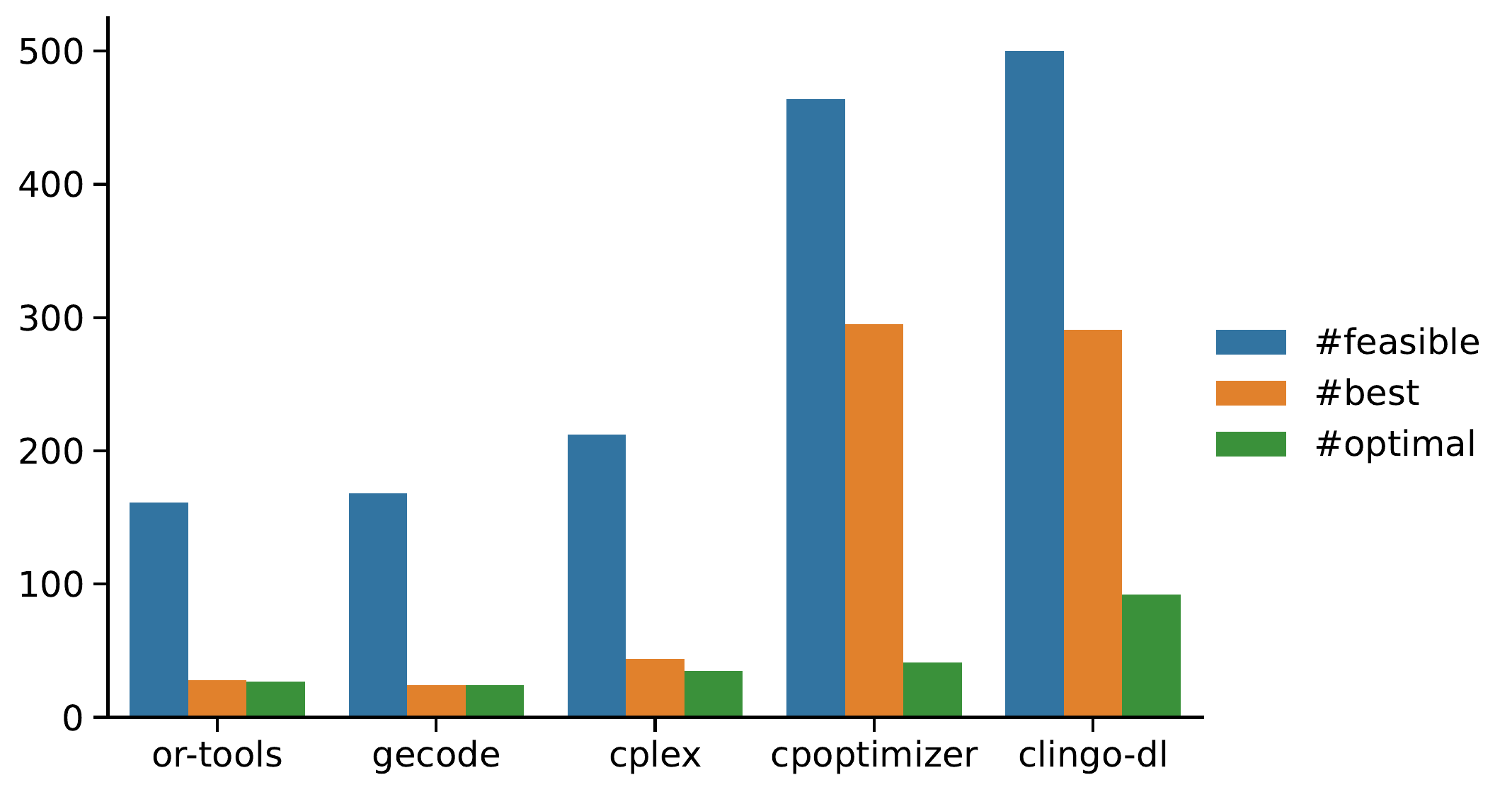}
    \end{subfigure}
    
    \centering
    \begin{subfigure}{\textwidth}\centering%
		\includegraphics[width=0.8\textwidth]{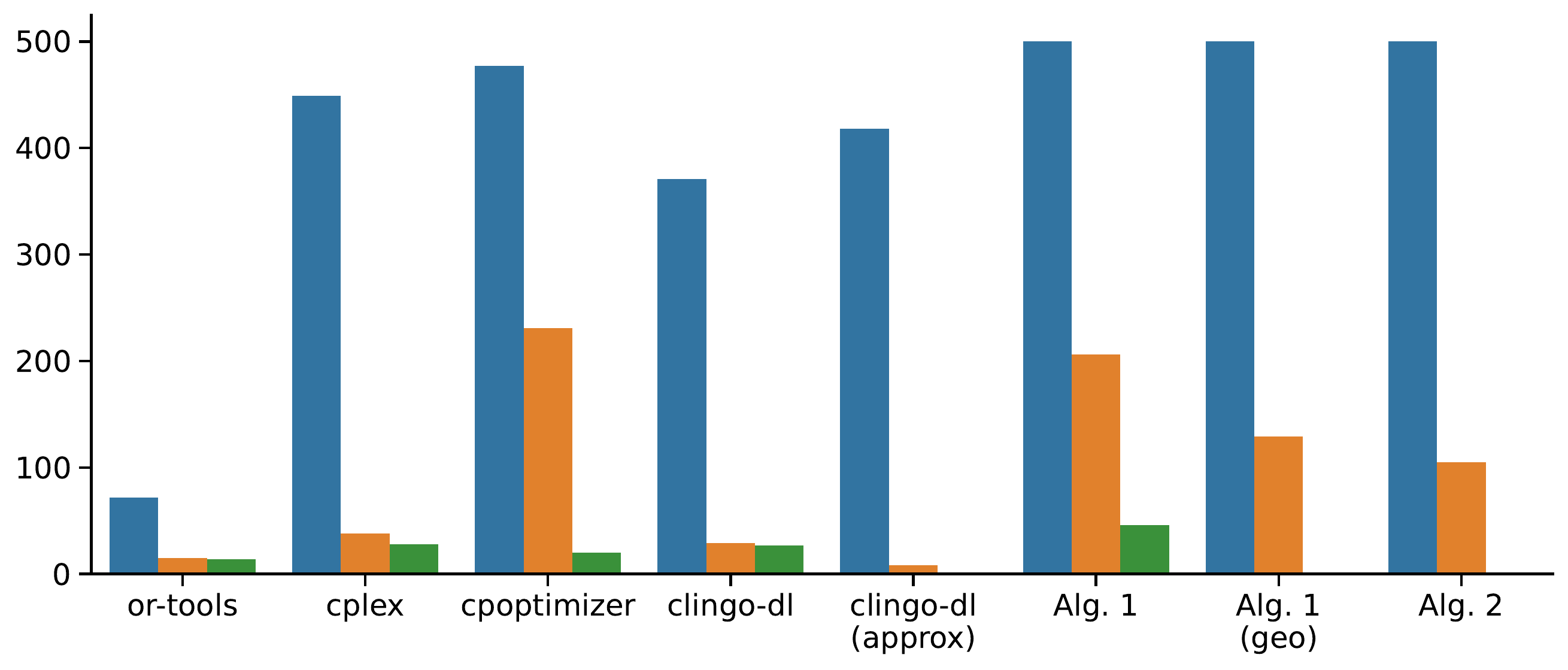}
    \end{subfigure}%
    \caption{Different systems on all instances for makespan (top) and lex-makespan (bottom) with a time limit of 15 minutes.}\label{fig:comparison-overall15}%
\end{figure}
%%%%%%%%%%%%%%%%%%%%%%%%%%%%%%%%%%%%%%%%%%%%%%%%%%%%%%%%%%%%%%%%%%%%%%%%%%%%%%%%%%%

Figures~\ref{fig:comparison-overall} and \ref{fig:comparison-overall15} give an overview of the performance of solvers
on all instances. We
compare 
the approaches for makespan and
lex-makespan optimisation .
For each solver, we report the number of instances where some solution was found (\emph{feasible}), a minimal solution among all approaches was found (\emph{best}), and a globally minimal solution was found (\emph{optimal}).

The makespan comparisons 
provide an important baseline as every approach that aims at improving lex-makespan necessarily
involves makespan minimisation, and any good lex-makespan optimisers also needs to produce small makespans.
With the practice oriented timeout of 5 minutes, the \clingoDL approach finds solutions for most of the instances and shows very good performance for the number of best and optimal solutions compared to \cplex, \cpoptimizer, \ortools and \gecode. At least when using our MiniZinc model, \ortools and \gecode 
have difficulties to find solutions for a large proportion of the instances. 
The performance of 
\cplex and \cpoptimizer is better, but it is still behind \clingoDL. It should be noted that we did not investigate further improvements for those solvers.
In the setting with 15 minutes run time, \cpoptimizer finds slightly more best solutions than \clingoDL which still solved the most instances to optimality. For the other solvers, the longer time out does not seem to make much difference. 

For the lex-makespan comparisons, 
we use \ortools, \cplex, \cpoptimizer 
as well as \clingoDL, the \clingoDL approximation described in Section~\ref{sec:approx} with 
granularity
$g=10$,  and Algs.~\ref{alg:exact} and \ref{alg:approx}. While \ortools, \cplex and \clingoDL struggle now to find feasible solutions in the 5 minutes setting, our multi-shot approaches shine in comparison with \cpoptimizer trailing closely behind. The \clingoDL approximation does find more feasible solutions than normal \clingoDL. However, it finds the least number of best solutions when compared to the others. While the approximation does indeed improve performance of \clingoDL on bigger instances, it performs worse on small instances and the results for the bigger instances are dwarfed by the other approaches.
If Alg.~\ref{alg:exact} is used without geometric timeouts, it reports optimal solutions for quite a number of instances and finds the most best solutions. This, when comparing with Algs.~\ref{alg:exact} and \ref{alg:approx} with geometric timeouts, is not surprising as more time, for many instances all the time, is spent on minimising the first component of the lex-makespan.
With the 15 minutes run time as shown in Figure~\ref{fig:comparison-overall15}, the results are generally similar. However, here \cpoptimizer manages to overtake Alg.~\ref{alg:exact} on the number of best solutions found even though the latter still proves more instances to be optimal. Furthermore, with the longer run time, \cplex does find significantly more feasible solutions. Overall, our ASP based solution is still largely competitive with the commercial solver \cpoptimizer.

Figures~\ref{fig:comparison-overall} and \ref{fig:comparison-overall15} do not tell us much about the quality differences of the schedules produced by the different algorithms.
As our informal objective is that machines complete as early as possible, 
we show in Figures~\ref{fig:dynamics} and Figures~\ref{fig:dynamics15} graphs for the ratio of machines that completed as a function 
of schedule time, i.e., $f(t) = M(S,t)/m$. 
This allows us to compare different approaches on the same instance classes, where the x-axis is schedule time in seconds and
the y-axis is the average of $M(S,t)/m$ over the instances.
The curves for different algorithms can be interpreted as follows: 
the earlier a curve reaches $1$ (all machines completed), the smaller is the average makespan of the  instances. The shape of the curve reveals details about the quality of the schedule prior to this point. For our informal objective, a steep incline of this curve is desired---the earlier it gets ahead and stays ahead, the better.

%%%%%%%%%%%%%%%%%%%%%%%%%%%%%%%%%%%%%%%%%%%%%%%%%%%%%%%%%%%%%%%%%%%%%%%%%%%%%%%%%%%
\begin{figure}[p]
    % \begin{subfigure}
            
    % \end{subfigure}
	\begin{tabular}{l@{}l@{}}%
			\input{plots/MStplot-CmaxVsLex.5min.5inrow.3.L.tex}	& \input{plots/MStplot-CmaxVsLex.5min.5inrow.3.H.tex}\\
			\input{plots/MStplot-CmaxVsLex.5min.5inrow.5.L.tex}	& \input{plots/MStplot-CmaxVsLex.5min.5inrow.5.H.tex}\\
			\input{plots/MStplot-CmaxVsLex.5min.5inrow.10.L.tex} & \input{plots/MStplot-CmaxVsLex.5min.5inrow.10.H.tex}\\
		    \input{plots/MStplot-CmaxVsLex.5min.5inrow.15.L.tex} & \input{plots/MStplot-CmaxVsLex.5min.5inrow.15.H.tex}\\
			\input{plots/MStplot-CmaxVsLex.5min.5inrow.20.L.tex} &
			\input{plots/MStplot-CmaxVsLex.5min.5inrow.20.H.tex}
	\end{tabular}
    \caption{Machine completion rate over time for makespan and lex-makespan (5 minutes run time).}\label{fig:dynamics}
\end{figure}
%%%%%%%%%%%%%%%%%%%%%%%%%%%%%%%%%%%%%%%%%%%%%%%%%%%%%%%%%%%%%%%%%%%%%%%%%%%%%%%%%%%

%%%%%%%%%%%%%%%%%%%%%%%%%%%%%%%%%%%%%%%%%%%%%%%%%%%%%%%%%%%%%%%%%%%%%%%%%%%%%%%%%%%

\begin{figure}[p]
    % \begin{subfigure}
            
    % \end{subfigure}
	\begin{tabular}{l@{}l@{}}%
			\input{plots/MStplot-CmaxVsLex.15min.5inrow.3.L.tex}	& \input{plots/MStplot-CmaxVsLex.15min.5inrow.3.H.tex}\\
			\input{plots/MStplot-CmaxVsLex.15min.5inrow.5.L.tex}	& \input{plots/MStplot-CmaxVsLex.15min.5inrow.5.H.tex}\\
			\input{plots/MStplot-CmaxVsLex.15min.5inrow.10.L.tex} & \input{plots/MStplot-CmaxVsLex.15min.5inrow.10.H.tex}\\
		    \input{plots/MStplot-CmaxVsLex.15min.5inrow.15.L.tex} & \input{plots/MStplot-CmaxVsLex.15min.5inrow.15.H.tex}\\
			\input{plots/MStplot-CmaxVsLex.15min.5inrow.20.L.tex} &
			\input{plots/MStplot-CmaxVsLex.15min.5inrow.20.H.tex}
	\end{tabular}
    \caption{Machine completion rate over time for makespan and lex-makespan (15 minutes run time).}\label{fig:dynamics15}
\end{figure}

We again start by discussing the results for the 5 minutes time limit.
We only consider Algs.~\ref{alg:exact} and~\ref{alg:approx} with geometric timeouts against plain makespan optimisation with \clingoDL in Figure~\ref{fig:dynamics}. We show instance classes of increasing size from left to right and compare instances of type ``low dedication'' in the upper row and ``high dedication'' in the lower row. 
All approaches produce small makespans and thus schedules with a high throughput.
This is worth emphasising since only half of the time limit is used here for makespan optimisation by Algs.~\ref{alg:exact} and~\ref{alg:approx}.
However, when comparing the shape of the curves, the lex-makespan optmisers show their strengths for meeting our informal objective; the difference to makespan is subtle for  Alg.~\ref{alg:exact} but more pronounced for Alg.~\ref{alg:approx}. While Alg.~\ref{alg:approx} finds fewer minimal solutions than Alg.~\ref{alg:exact} according to Figure~\ref{fig:comparison-overall}, it tends to get ahead the earliest in terms of completed machines when considering the execution of the schedules, especially for high dedication instances. We can quantify this by looking at the average ratio of machines finished at each point in time. On average, Alg 1. improves this metric by 6.3\% whereas Alg. 2 shows an improvement of 9.53\%.

The graphs for the longer run time of 15 minutes shown in Figure~\ref{fig:dynamics15} are largely similar. However, the improvements of Alg.~\ref{alg:approx} are slightly more pronounced, especially for the instances with 15 machines.

% summary
In summary, \clingoDL performs best among all considered approaches for plain makespan optimisation. For lex-makespan, Alg.~\ref{alg:exact} performs best when the time limit is 5 minutes, while \cpoptimizer finds more often better solutions with the longer time limit of 15 minutes.

%%%%%%%%%%%%%%%%%%%%%%%%%%%%%%%%%%%%%%%%%%%%%%%%%%%%%%%%%%%%%%%%%%%%%%%%%%%%%%%%%%%%%
\section{Related Work}\label{sec:rel}
%%%%%%%%%%%%%%%%%%%%%%%%%%%%%%%%%%%%%%%%%%%%%%%%%%%%%%%%%%%%%%%%%%%%%%%%%%%%%%%%%%%%%

\subsection{Parallel Machine Scheduling}
Many variants of Parallel Machine Scheduling Problem, e.g., \citep{allahverdi2008,allahverdi2015third}, have been studied extensively in the literature. Previous publications have considered eligibility of machines, e.g., \citep{afzalirad2016,Perez-Gonzalez2019,Bektur2019}, machine dependent processing time, e.g., \citep{Vallada2011,Avalos-Rosales2013,allahverdi2015third}, and sequence dependent setup times, e.g., \citep{Vallada2011,Perez-Gonzalez2019,Fanjul-Peyro2019,Gedik2018}.

\subsection{Lexicographical Makespan}
The idea to use 
lexicographical makespan optimisation to obtain robust schedules for identical 
parallel machines comes from \cite{letsios2021exact} but has not been used, to the best of our knowledge, when setup times are present. 
The general idea of optimising not only the element that causes the highest costs but also
the second one and so on to obtain robustness, fairness, or balancedness is
studied under the notion of min-max optimisation
for a various combinatorial problems~\citep{burkard1991lexicographic,ogryczak2006direct}.

There are several related objective functions that can be used to achieve similar effects as minimising the lex-makespan. 
Load balancing can be used to obtain balanced resource utilisation by equalising the workload on machines~\citep{rajakumar2004workflow,yildirim2007parallel,Sabuncu20}. 
One measure for this is to minimise the relative
percentage of imbalance in workload~\citep{rajakumar2004workflow} which is determined based on the difference between a machine span and the makespan.
Other approaches try to minimise the difference between the largest and the smallest machine span~\citep{ouazene2014workload}.
Note that the machines $m_2$ and $m_3$ in Figure~\ref{fig:schedules} (a) are balanced under this notion, 
but this does not 
ensure that the machines finish as early as they could. 

\subsection{ASP Applications}
\cite{Sabuncu20} provide a novel formulation of a machine scheduling problem in ASP. Their approach bears some similarity to ours, but their objective is to balance the workload of the given machines.
In general, load balancing can lead to schedules where, e.g., longer than necessary setup  times  are used to artificially prolong machine spans
for reducing imbalances. 
Another idea is to minimise workload instead of balancing it. While this achieves short processing times, it does not ensure that all machines  finish as early as possible either. 
Similar to the workload, minimising the sum of machine spans does not prevent that jobs are scheduled in an unbalanced way;  Figs.~\ref{fig:schedules} (b) and (c) serve as an example of two schedules with the same total machine span. 
Minimising a non-linear sum of machine spans like their squares comes close to our informal objective but is different from the lex-makespan
as it does not guarantee a minimal makespan~\citep{walter2017note}. 
Another way to obtain compact schedules is it to
minimise the total completion times~\citep{weng2001unrelated}.
This however can  pull short jobs to the front of the schedule, which 
can 
adversely interfere with avoiding large setup times.

\subsection{ASP Extensions}
Extending ASP with difference logic is just one way to blend integer constraints and ASP and
there are several other approaches~\citep{lierler2014relating,gebser2009constraint}.
We evaluated ASP with full integer constraints with
\clingcon, but performance on our problem was poor. The fast propagation enabled by the lower computational complexity of difference logic seems a big advantage here.
The \clingoDL system has indeed been used for the related problem of job-shop scheduling and 
makespan optimisation~\citep{janhunen2017clingo,kholany}. 
Train scheduling for the Swiss Federal Railways is another application 
of \clingoDL 
that involves  routing, scheduling,  and 
complex optimisation~\citep{abels2019train}. 
However, we are solving a different problem with a more complex objective function and also compare to other solving paradigms.

%%%%%%%%%%%%%%%%%%%%%%%%%%%%%%%%%%%%%%%%%%%%%%%%%%%%%%%%%%%%%%%%%%%%%%%%%%%%%%%%%%%%%
\section{Conclusion}\label{sec:concl}
%%%%%%%%%%%%%%%%%%%%%%%%%%%%%%%%%%%%%%%%%%%%%%%%%%%%%%%%%%%%%%%%%%%%%%%%%%%%%%%%%%%%%

We studied the application of hybrid ASP with difference logic to solve a challenging parallel machine scheduling problem with setup-times in industrial semi-conductor production at Bosch. 
As objective function, we used the lex-makespan which generalises the canonical makespan to a tuple of machine spans and aims at accomplishing short completion times for all machines.
Semi-conductor production involves not only one but several connected work centers that solve similar problems. Having a flexible ASP solution for one that can easily be adapted to others is highly desirable. 
To make ASP perform up to par,
we appropriated advanced techniques like difference constraints, 
multi-shot solving, domain heuristics, and approximations for our application.

For  the experimental evaluation, we considered random instances of realistic size and structure. We further implemented a solver-independent MiniZinc model as well as direct encodings that we used for comparisons with MIP and CP solvers.
The results show that the objective of short completion times for machines
is well achieved.
Performance is improved by using approximations without significant deterioration of the  schedules produced. 
It is encouraging that the ASP approaches turn out to be competitive with commercial MIP and CP solvers which are, at least to some extent, engineered for industrial scheduling problems.
 
We plan to study meta-heuristics for lex-makespan optimisation  in combination with ASP and methods to combine the lex-makespan with other common objective functions.

%%%%%%%%%%%%%%%%%%%%%%%%%%%%%%%%%%%%%%%%%%%%%%%%%%%%%%%%%%%%%%%%%%%%%%%%%%%%%%%%%%%%
\section*{Acknowledgments}
%%%%%%%%%%%%%%%%%%%%%%%%%%%%%%%%%%%%%%%%%%%%%%%%%%%%%%%%%%%%%%%%%%%%%%%%%%%%%%%%%%%%
We would like to thank the reviewers for their comments and Michel Janus, Andrej Gisbrecht, and Sebastian Bayer for very helpful discussions on the scheduling problems at Bosch, as well as 
Roland Kaminski,
Max Ostrowski,
Torsten Schaub, and
Philipp Wanko for their help, valuable suggestions, and feedback on constraint ASP. Johannes Oetsch was supported by funding from the Bosch Center for Artificial Intelligence.

\end{document}